\documentclass[11pt,a4paper]{article}
\usepackage[hyperref]{naaclhlt2019}
\usepackage{times}
\usepackage{latexsym}
\usepackage[utf8]{inputenc}
\usepackage[arabic,USenglish]{babel}
\usepackage{enumitem}
\usepackage{float}
\usepackage{url}

\aclfinalcopy % Uncomment this line for the final submission
\title{Leveraging Pretrained Word Embeddings for Part-of-Speech Tagging of Code Switching Data}

\author{Fahad AlGhamdi $^1$, and Mona Diab$^1$$^,$$^2$ \\
 $^1$ Department of Computer Science, The George Washington University  \\
 $^2$ AWS, Amazon AI \\
  {\tt \{fghamdi, mtdiab\}@gwu.edu} }

\date{}

\begin{document}
\maketitle
\begin{abstract}

Linguistic Code Switching (CS) is a phenomenon that occurs when multilingual speakers alternate between two or more languages/dialects within a single conversation. Processing CS data is especially challenging in intra-sentential data given state-of-the-art monolingual NLP technologies since such technologies are geared toward the processing of one language at a time. In this paper, we address the problem of Part-of-Speech tagging (POS) in the context of linguistic code switching (CS).  We explore leveraging multiple neural network architectures to measure the impact of different pre-trained embeddings methods on POS tagging CS data. We investigate the landscape in four CS language pairs, Spanish-English, Hindi-English, Modern Standard Arabic- Egyptian Arabic dialect (MSA-EGY), and Modern Standard Arabic- Levantine Arabic dialect (MSA-LEV). Our results show that multilingual embedding (e.g., MSA-EGY and MSA-LEV) helps closely related languages (EGY/LEV) but adds noise to the languages that are distant (SPA/HIN). Finally, we show that our proposed models outperform state-of-the-art CS taggers for MSA-EGY language pair.

\end{abstract}
\section{Introduction}
Code Switching (CS) is a common linguistic behavior where two or more languages/dialects are used interchangeably in either spoken or written form. CS is typically present at various levels of linguistic structure: across sentence boundaries (i.e., inter-sentential), within the same utterances, mixing two or more languages (i.e., intra-sentential), or at the words/morphemes level. The CS phenomenon is noticeable and common in countries that have large immigrant groups, naturally leading to bilingualism. Typically people who code switch master two (or more) languages: a common first language (lang1) and another prevalent language as a second language (lang2). The languages could be completely distinct such as Mandarin and English, or Hindi and English, or they can be variants of one another such as in the case of Modern Standard Arabic (MSA) and Arabic regional dialects (e.g. Egyptian dialect-EGY). CS is traditionally prevalent in the spoken modality but with the ubiquity of the Internet and proliferation of social media, CS is becoming ubiquitous in written modalities and genres \cite{HindiEnglshiCS,TheLanguageBook,CSANDMC}. This new situation has created an unusual deluge of CS textual data on the Internet. This data brings in its wake new opportunities, but it poses serious challenges for different NLP tasks; traditional techniques trained for one language tend to break down when the input text happens to include two or more languages. The performance of NLP models that are currently expected to yield good results (e.g., Part-of-Speech Tagging) would degrade at a rate proportional to the amount and level of mixed-language present. %The performance of the state-of-the-art monolingual POS taggers degrade dramatically when the input text includes two or more languages. 
This is a result of out-of-vocabulary words in one language and new hybrid grammar structures, and in some cases shared cognates or ambiguous words that exist in both language lexicons.

%POS tagging is the task where each word in text is contextually labeled with grammatical labels such as, noun, verb, proposition, adjective, etc. 

%%higher-up >>> More adavnced NLP tools.
POS tagging is a vital component of any Natural Language Understanding system, and one of the first tasks researchers employ to process data. POS tagging is an enabling technology
needed by higher-up NLP tools such as chunkers and parsers – syntactic, semantic and discourse level processing; all of which are used for such applications as sentiment analysis and subjectivity, text summarization, information extraction, automatic speech recognition, and machine translation among others. As such, it is crucial that POS taggers be able to process CS textual data. 

In this paper, we address the problem of Part-of-Speech tagging (POS) for CS data on the intra-sentential level for multiple language pairs. We explore the effect of using various embeddings setups and multiple neural network architectures in order to mitigate the problem of the scarcity of CS annotated data. We propose multiple word embedding techniques that could help in tackling POS tagging of CS data.

In order to examine the generalization of our approaches across language pairs, we conduct our study on four different evaluation CS data sets, covering four language pairs: Modern Standard Arabic and the Egyptian Arabic dialect (MSA-EGY), Modern Standard Arabic- and the Levantine Arabic dialect (MSA-LEV), Spanish-English (SPA-ENG) and Hindi-English (HIN-ENG). We use the same POS tag sets for all language pairs, namely, the  Universal POS tag set \cite{UDPOS}. Our contributions are the following: a) We use a state-of-the-art bidirectional recurrent neural networks; b) We explore different strategies to leverage raw textual resources for creating pre-trained embeddings for POS tagging CS data; c) We examine the effect of language identifiers  for joint POS tagging and language identification; d) We present the first empirical evaluation on POS tagging with four different language pairs. All of the previous work focused on a single or two language pair combinations.

\section{Related Work}
Developing CS text processing NLP techniques for analyzing user generated content as well as cater for needs of multilingual societies is vital \cite{Vyas-2015}. Previous studies that address the problem of POS tagging  of CS data first attempt to identify the correct language of a word before feeding it into an appropriate monolingual tagger \cite{solorio2008,HindiEnglshiCS,MyPOSPaper}. As is typically the case in NLP, such pipelines suffer from the problem of error propagation; e.g., failure of the language identification will cause problems in the POS tag prediction. 
Other approaches have trained supervised models on POS-annotated, CS data resources which are expensive to create and unavailable for most language pairs. \cite{MyPOSPaper,solorio2008,jamatia2015,W16-5804} 
\newcite{solorio2008}, proposed the first statistical approach to POS tagging of CS data where they employ several heuristics to combine monolingual taggers with limited success, achieving 86\% accuracy when choosing the output of a monolingual tagger based on the dictionary language ID of each token. However, an SVM trained on the output of the monolingual taggers performed better than their oracle, reaching 93.48\% accuracy. 
%\citep{solorio2008} presents a machine learning based model that outperforms all baselines on SPA-ENG CS data. Their system  utilizes only a few heuristics in addition to the monolingual taggers. 
\newcite{sequierapos} introduces a ML-based approach with a number of new features for HIN-ENG POS tagging for Twitter and Facebook chat messages. The new feature set considers the transliteration problem inherent in social media. Their system achieves an accuracy of 84\%. \newcite{jamatia2015} use both a fine-grained and coarse-grained POS tag set in their study. They introduce a comparison between the performance of a combination of language specific taggers and a machine learning based approach that uses a range of different features. They conclude that the machine learning approach failed to outperform the language specific combination taggers.
%our paper
\newcite{MyPOSPaper} examine seven POS tagging strategies to leverage the available monolingual resources for CS data. They conducted their study on two language pairs, namely MSA-EGY and SPA-ENG. The proposed strategies are divided into combined conditions and integrated conditions.  Three of the combined conditions consist of running monolingual POS taggers and language ID taggers in a different order and combining the outputs in a single multilingual prediction. The fourth combined condition involved training an SVM model using the output of the monolingual taggers. The three integrated approaches trained a monolingual state-of-the-art POS tagger on a) CS corpus; b) the union of two monolingual corpora of the languages included in the CS; c) the union of the corpora used in a and b. Both combined and integrated conditions outperformed the baseline systems. The SVM approach consistently outperformed the integrated approaches achieving the highest accuracy results for both language pairs. 
%victor
\newcite{VictorPaper} propose a new approach to POS tagging for code switching SPA-ENG language pair based on recurrent neural network (RNN) as a way of providing better tools for better code switching data processing, including POS taggers.  The authors use an experimental approach of POS tagging for CS utterances entailing use of state-of-the-art bi-directional RNN to extensively study effects of language identifiers. For the Boolean features, a bi-directional Long short-term memory (BiLSTM) state-of-the-art neural network with suffix, prefix, and word embeddings, the results show that the neural POS tagging model proposed by the authors performs comparatively higher than other state of the art CS taggers, their system yields a POS tagging accuracy of  96.34\%, while joint POS and language ID tagging yields an accuracy of 96.39\% for POS tagging and language ID accuracy of 98.78\% \cite{VictorPaper}. 
 Language Modeling (LM) on HIN-ENG language texts has been proposed by \cite{W18-3211}. \cite{D18-1344}  propose the use of a computational technique to create artificial code mixed data that is grammatically valid based on the ECT (Equivalence Constraint Theory) to solve the challenge of scarcity of CS training language models using an experimental approach.
%using two bilingual embedding techniques
 \cite{D18-1344} uses two bilingual embedding techniques, namely Bilingual Compositional Model (BiCVM) and Correlation Based Embeddings (BiCCA) \cite{faruqui2014, blunsom2014}. Word embedding results in improved semantic and syntactic CS processing tasks. BiCVM at the sentence level only yields better performance for semantic tasks. BiCCA also only do well on semantic tasks because they make use of word alignments. Furthermore, g-skip embedding outperformed BiCCA and BiCVM, performing well across syntactic and semantic tasks. The study by \newcite{D18-1344} illustrates that using pretrained embeddings learned from CS data outperforms pretrained embeddings learned from standard bilingual embeddings.
 %Hinid state of the art system.
\newcite{bhat2018universal} introduce a dependency parser developed specifically for HIN-ENG CS data. They adopted the neural stacking architecture proposed by \cite{weiss,chen} for learning POS tagging and parsing and for transferring the knowledge from bilingual models trained on Hindi and English UD treebanks. The stack-prop tagger, the model proposed by \cite{weiss}, achieves the highest accuracy of 90.53. To identify the language ID, they train a multilayer perceptron (MLP) stacked on top of a recurrent bidirectional LSTM (Bi-LSTM) network. The results of their system is 97.39 \%. 
 % by leveraging POS information from Bilingual tagger using neural stacking. 
\section{Approach}
Pretrained word embeddings enable models to exploit the raw textual data which is in all languages larger than annotated data. In recent times, there has been some interest in embedding approaches, e.g., multilingual embeddings and bilingual word embeddings, where two monolingual embeddings of two languages are mapped to a shared embeddings space. The main advantage of bilingual and multilingual embeddings is in solving tasks involving reasoning across two languages, such as Machine Translation (MT) \cite{vulic2016,zou2013}, as well as enabling transfer of models learned on a resource-rich language onto a resource-poor language \cite{adams2017}. One of the potential applications of bilingual and multilingual embeddings is in the processing of code switching language. In this section, we compare leveraging multiple neural network architectures for POS tagging CS data.
Moreover, we explore different embedding setups to investigate the optimal way of tackling the POS tagging of CS data. First, we illustrate the tool used for training the embeddings layer.  Second, we present the neural network models. Then we list the embedding setups. 
%%%%%%%%%
\paragraph{Pretrained Word Embedding Model} 
Most successful methods for learning word embeddings \cite{mikolov2013,glove,fasttext} rely on the distributional hypothesis \cite{mikolov}, i.e., words occurring in similar contexts tend to have similar meanings. 
Among all word embedding techniques, we choose the FastText tool developed by Facebook \cite{fasttext}. Our choice of FastText is motivated by the fact that social media networks are the primary source of our unlabeled data. This type of data exhibits huge variations of spelling and misspellings. FastText takes advantage of subword (i.e., n-gram) information. It creates vector representations for misspelling replacement candidates absent from the trained embedding space, by summing the vectors of the character n-grams. All word embedding techniques aim to capture the relation of the words that tend to appear in similar context. These relations occur on the sentential level for most of the NLP applications trained for monolingual data. However, the lack of having sufficient context in CS data makes learning these kinds of relations a difficult task as they occur on the sub-sentential (intra-sentential) level. Hence, we resort to using FastText due to its principle approach of leveraging subword (i.e., n-gram) information. The n-gram approach resolves the problem of modeling languages with rare word inflections by using character n-grams. On the other hand, other embedding techniques, e.g., word2vec and glove, lack subword information and hence struggle with morphologically rich languages such as Arabic and noisy data such as Twitter data \cite{word2vec, glove}.

\paragraph{A Model for Neural POS Tagging} 

%For our experiments we use two variants of BiLSTM architectures similar to the onse  proposed by \cite{bilstm} for POS tagging, one for joint POS tagging for related language pairs, and one for joint POS and Language ID tagging. 
For our experiments we use three neural networks architectures: a) a BiLSTM-CRF architecture similar to the one proposed by \cite{bilstm} for POS tagging; b) a multi-task learning model for learning jointly POS tagging for related language pairs; and, c) a multi-task learning model for learning jointly POS and Language ID tagging.

%In the most basic architecture, we initialize the embedding layer with the pre-trained FastText word embeddings and feed the output sequence from this layer to the BiLSTM layer. At the output layer a softmax activation function is applied over the hidden representation learned in the BiLSTM layer. The BiLSTM hidden layer has 200 units for each direction and dropout of 0.2. We use early stopping \cite{keras} based on performance on validation sets. According to our experiments, the best parameters appear at around 50 epochs, according to our experiments. We refer to this model as BiLSTM POS Tagger for the rest of the article and in our tables.
%%%%%%%%%%%%%%%%%%%%%%%%%%%%%%%%%%%%%%%%%%%%%%%%%%%%
%%%%%%%%%%%%%%%   Review  %%%%%%%%%%%%%%%%%%%%%%%%%%
%%%%%%%%%%%%%%%%%%%%%%%%%%%%%%%%%%%%%%%%%%%%%%%%%%%%
\begin{itemize}
\item {We train a BiLSTM network with a conditional random field objective \cite{bilstm} that obtain the probability distribution over all labels by jointly modeling the probability of the entire tag sequence score. We initialize the embedding layer with the pre-trained FastText word embeddings and feed the output sequence from this layer to the BiLSTM layer. The BiLSTM hidden layer has 200 units for each direction and dropout of 0.2. We use early stopping \cite{keras} based on performance on validation sets. We refer to this model as BiLSTM-CRF POS Tagger for the rest of the article and in our tables.
}
\item {Our second model is a multi-task learning model that learns simultaneously POS tagging for related code switching language pairs. The architecture of this model follows the BiLSTM-CRF architecture \cite{bilstm}. We add a second CRF output layer for predicting the POS tag for the related language pair (e.g., MSA-LEV), while the first CRF output layer is for predicting the POS tag for the other related language pair (e.g., MSA-EGY). The two output layers of POS tagging tasks for the two language pairs are independent and are connected by their weight matrices to the hidden layer, and both loss functions are given the same weight. We refer to this model as MTL-POS Tagger for the rest of the article and in our tables.}

\item{The third model is a multi-task learning model. The model learns simultaneously POS and language id tags with the aim of boosting the performance of POS tagging task. The architecture of this model follows MTL-POS Tagger architecture. The difference is that one output layer is for predicting the POS tagging, and the other output layer is for predicting language id task. This model is referred to as MTL-POS+LID Tagger.}
\end{itemize}

\paragraph{Experimental Conditions}

\paragraph{Monolingual embedding (baseline):}
We train word embeddings using the monolingual corpora for each language involved in the four language pairs. The results of this approach are six separate pre-trained embeddings,  MSA, EGY, LEV, ENG, SPA, and HIN pre-trained embeddings. For each language, we train a BiLSTM-CRF model using one of the six pre-trained embeddings. We consider these models as baseline systems. The baseline performance is the POS tagging accuracy of the monolingual models with no special training for CS data.
%%%%%%%%%%%%%%%%%%%%%%%%%%%%%%%%
\paragraph{Merged Bilingual embeddings:}
%%%%%%%%%%%%%%%%%%%%%%%%%%%%%%%%
\paragraph{Combined filtered monolingual corpora (CFM):} To leverage the inter-sentential code switching type, we train a model using a pre-trained word embedding trained on monolingual data only. The assumption is that the data in MSA is purely MSA and that in EGY is purely EGY. This condition yields an inter-sentential CS pre-trained embeddings. None of the sentences reflect intra-sentential CS data.
%%%%%%%%%%%%%%%%%%%%%%%%%%%%%%%%
\paragraph{Pure Code switched corpora (PCS):}
To leverage the intra-sentential code switching type, we train a model using a pre-trained word embedding trained on CS data only. The assumption is that the data used to train the embeddings exhibit the CS phenomenon. This condition yields an intra-sentential CS type. None of the sentences reflect inter-sentential CS data.
%%%%%%%%%%%%%%%%%%%%%%%%%%%%%%%%
\paragraph{(Pseudo) Combined monolingual and CS corpora (PseudoCS):} To address both code switching types, the intra-sentential and inter-sentential, we combine the pure code switched corpora and combined filtered monolingual corpora to form unified to be used to train a unified word embeddings.
%%%%%%%%%%%%%%%%%%%%%%%%%%%%%%%%RRRR
\begin{table}[H]
\centering
\renewcommand{\arraystretch}{1}
\setlength{\tabcolsep}{0.4em}
\small
\begin{tabular}{|c|c|c|c|}
\hline
\textbf{Dataset}   & \textbf{Train} & \textbf{Dev} &  \textbf{Test} \\ \hline
\textbf{ARZ (MSA-EGY)}       &  133,357             &     21,146     &    20,464   \\ \hline
\textbf{LEV (MSA-LEV)} &        45,167                & 5,749   &         5,779 \\ \hline
\textbf{Miami Bangor (SPA-ENG)}    & 268,464              &   67,114   &   67,114   \\ \hline
\textbf{UD-HIN-ENG}    & 19,695              &   3,339   &   3,190   \\ \hline

\end{tabular}
\caption{Datasets distribution for the four language pairs}
\label{DatasetDist}
\end{table}
%%%%%%%%%%%%%%%%%%%%%%%%%%%%%%%%

\paragraph{Merged multilingual embeddings:}
We train a multilingual embedding for language pairs that have one language in common (pivot language).
To do so, we combine the corpora used to train the word embeddings of the language pairs that share one common language. For MSA-EGY and MSA-LEV language pairs, we have a common language, which is MSA, while ENG is a common language between SPA-ENG and HIN-ENG. To leverage this commonality between each of the two language pairs, we merge all the previous corpora: PCS, Mono, and PseudoCS for each language pair to form one corpus used to train merged multilingual embeddings model for MSA-EGY-LEV languages and another corpus for SPA-HIN-ENG. The intuition of this embeddings is to capture word usage in the context of each language and eliminates the ambiguity for the words that have the same surface form in multiple languages. 

\paragraph{Projected bilingual embedding}
%Recently, there has been interest in learning bilingual representations. 
Projected bilingual embeddings are vector representations of two languages mapped into shared space, such that translated word pairs have similar vectors. There are three approaches to learn bilingual embeddings: 1) by mapping the space of both monolingual embeddings into a single shared space; 2) monolingual adaptation of one language’s embedding space into another's; 3) Bilingual Training by bootstrapping the target representations learned from well-trained embeddings space of a source language. 
We train individual CS and monolingual embedding models separately before mapping them into a shared embedding space. To do so, We use MUSE \cite{muse}, state-of-the-art model for creating a projected bilingual embedding that uses the monolingual adaptation technique to create the shared space embedding. MUSE is equipped to learn either via supervision or no supervision. In our study, we utilize the unsupervised version of MUSE.  
%The results of applying MUSE on our data are six shared space embeddings. The embeddings are Bi-MSA-EGY, Bi-MSA-LEV,  Bi-SPA-ENG, and Bi-HIN-ENG. 
\section{Evaluation}
%%%%%%%%%%%%%%%%%%%%%%%%%%%%%%%%

\subsection{Data sets}

%Throughout our experiments, we use one evaluation dataset for each language pair and various corpora for training the embeddings layer. In this section, we describe these data sets. Data sets distribution is shown in Table-\ref{DatasetDist}.
Throughout our experiments, we use one evaluation dataset for each language pair and various corpora for training the embeddings layer. Table-\ref{DatasetDist} shows the distribution of the evaluation data sets.

%The Miami Bangor corpus, which contains instances of inter-sentential and intra-sentential CS utterances in SPA-ENG, is used for training and testing SPA-ENG CS models and comparing these to monolingual models. 
%The LDC Egyptian Arabic Treebanks 1-5 (ARZ1-5) (Maamouri et al., 2012). The ARZ1-5 data is from the discussion forums genre mostly in the Egyptian Arabic dialect (EGY). It contains instances of inter-sentential and intra-sentential CS utterances in SPA-ENG, is used for training and testing MSA-EGY CS models and comparing these to monolingual models. 
%Curras dataset, which includes instances of inter-sentential and intra-sentential CS utterances in MSA-LEV, is used for training and testing MSA-LEV CS models and comparing these to monolingual models. 
%Finally, the Universal Dependencies (UD) corpus, which contains instances of inter-sentential and intra-sentential CS utterances in HIN-ENG, is used for training and testing HIN-ENG CS models and comparing these to monolingual models. 
\paragraph{MSA-EGY}
%while the number of words is 175,367 words
%Each part of the ARZ data set is divided into train, test and dev sets.
We use the LDC Egyptian Arabic Treebanks 1-5 (ARZ1-5) \cite{Maamouri2012}. The ARZ1-5 data is from the discussion forums genre mostly in the Egyptian Arabic dialect (EGY). The total number of sentences in the corpus is 13,698 while the number of words is 174,967 words. 

To train pre-trained embeddings, we crawl Egyptian tweets from some of the Egyptian public figures' Twitter accounts. The rest of the Egyptian raw textual data comes from the following sources: \cite{zaidan2011arabic}'s Egyptian online news commentary corpus, the Egyptian tweets used in the CS shared tasks \cite{SolorioEtAl:14,molina,cs-shared-tasl18}, portion of MSA Gigaword \cite{parker2011}, and LDC Arabic tree bank corpora (MSA) \cite{maamouri2004penn,diab2013ldc}. To identify the language id of MSA-EGY tokens, we use the Automatic Identification of Dialectal Arabic (AIDA2) tool \cite{AIDA2} to perform token level language identification for the EGY and MSA tokens in context.
%Review: Add the ATB CITE + Gigaword CITE

\paragraph{MSA-LEV}
We use the Curras Corpus of Palestinian Arabic as the MSA-LEV textual CS data \cite{curras}. Palestinian Arabic is a sub-dialect of Levantine Arabic. The corpus comprises  57,000 words, half of which come from transcripts of a TV show and the rest of which comes from various sources such as Facebook, discussing forums, Twitter, and blogs. The corpus is morphologically annotated by \newcite{Lev-dataset} using the same guidelines utilized for annotating the Egyptian ARZ corpus. We annotate the MSA-LEV evaluation data set with language id using the guidelines and tool proposed by \cite{CSPaper,wasa}.

To train pre-trained embeddings, we crawl tweets from Levantine public figures. Moreover, we compile Levantine and MSA raw textual data from multiple resources:  online news commentary corpus from \cite{zaidan2011arabic}, weblogs from COLABA \cite{diab2010colaba}, commentaries and tweets from \newcite{Ryan-dataset}, the Levantine portion of the PADIC data set \cite{PADIC-dataset}, portion of MSA Gigaword \cite{parker2011}, and LDC Arabic tree bank corpora (MSA) \cite{maamouri2004penn,diab2013ldc}.
%Review: Add the ATB CITE + Gigaword CITE

\paragraph{SPA-ENG}
The Miami Bangor (MB) corpus is a conversational speech corpus recorded from bilingual Spanish-English speakers living in Miami, FL. It includes 56 conversations recorded from 84 speakers \cite{soto2017}. The corpus consists of 242,475 words (333,069 including punctuation tokens) and 35 hours of recorded conversation. The language markers in the corpus were manually annotated. 

To train pre-trained embeddings for Spanish and English language, we use the English Universal Dependencies (UD) corpus \cite{L14-1067} and the Spanish UD corpus \cite{P13-2017}. Universal Dependencies (UD) is a project to develop cross-linguistically consistent treebank annotations for many languages. Moreover, we use some other English monolingual data from various resources. The English monolingual data contains around 250M sentences.

\paragraph{HIN-ENG}
The Hindi-English Code switching treebank is based on CS tweets of Hindi and English multilingual speakers (mostly Indian) \cite{Eng-Hin-CS-data}. % on Twitter HIN-ENG CS data \cite{Eng-Hin-CS-data}. 
The treebank is manually annotated using UD scheme. %The training and evaluations sets were separately annotated by different annotators using UD v2.  
The corpus contains data from Twitter. The corpus contains 1,852 tweets and 26,224 tokens. 
To train pre-trained embeddings for Hindi and English language, we use the English Universal Dependencies (UD) corpus \cite{L14-1067} and the Hindi UD corpus \cite{bhathindi,palmer2009hindi}. Also, we use some other Hindi monolingual data from various resources. The word representations are learned using Skip-gram model with negative sampling which is implemented in FastText toolkit for all language.  

%The training and evaluations sets were separately annotated by different annotators using UD v2 and v1 guidelines respectively

%%%%%%%%%%%%%%%%%%%%%Baseline Systems%%%%%%%%%%%%%%%%%%%%%%%%%%%%%%%%%%%
\paragraph{Baseline Results} The baseline performance is the POS tagging accuracy of the monolingual models with no special training for CS data. Therefore, our baselines are the neural network models trained using the monolingual embeddings. If CS data do not pose any particular challenge to monolingual POS taggers, then we should not expect a major degradation in performance. Table-\ref{Results-Table} shows the performance of the different baseline POS tagging systems on the test data. For each language pair, there are two baseline systems. 

For MSA-EGY, the baseline accuracies are 85.40 when the baseline system utilizes the MSA pre-trained embeddings, and 81.06 when BiLSTM-CRF uses the EGY pre-trained embeddings. Similarly, we report the baseline results for MSA-LEV, SPA-ENG and HIN-ENG language pairs.

%The accuracy results of MSA-LEV baseline systems are 83.25 when using MSA pre-trained embeddings, and 80.21 when the baseline system utilizes the LEV pre-trained embeddings. 
%For the SPA-ENG, the accuracy results of the baseline systems are 50.30 using SPA pre-trained embeddings and 71.21 using ENG pre-trained embeddings. 
%Finally, the accuracy results of HIN-ENG are 63.30 using HIN pre-trained embeddings and 67.11 suing ENG pre-trained embeddings.
%when the word pre-trained embeddings used by BiLSTM-CRF are HIN pre-trained embeddings, and ENG pre-trained embeddings.
%%%%%%%%%%%%%%%%%%%%%%%%%%%%%%%%%%%%%%%
%%%%%%%%%%%%%%%%%%%%%%%%%%%%%%%%%%%%%%%
\makeatletter
\def\thickhline{%
  \noalign{\ifnum0=`}\fi\hrule \@height \thickarrayrulewidth \futurelet
   \reserved@a\@xthickhline}
\def\@xthickhline{\ifx\reserved@a\thickhline
               \vskip\doublerulesep
               \vskip-\thickarrayrulewidth
             \fi
      \ifnum0=`{\fi}}
\makeatother

\newlength{\thickarrayrulewidth}
\setlength{\thickarrayrulewidth}{3\arrayrulewidth}

%%%%%%%%%%%%%%%%%%%%%%%%%%%%%%%%%%%%%%%
%%%%%%%%%%%%%%%%%%%%%%%%%%%%%%%%%%%%%%%

\begin{table*}[t!]
\small
\center
\renewcommand{\arraystretch}{.95} %to adjust the height of the cells (compress-row)
\begin{tabular}{|c|c|c|c|c|}
\hline
\textbf{Embedding Condition} & \textbf{MSA-EGY} & \textbf{MSA-LEV} & \textbf{SPA-ENG} & \textbf{HIN-ENG} \\ \hline
BiLSTM-CRF Tagger (1) + Random-Initi-Embed & 89.12 & 88.90  & 95.33  & 69.87  \\ \hline
(1) + mono (MSA/SPA/HIN) (Baseline) & 85.40 & 83.25 & 50.30 & 63.30 \\ \hline
(1) + mono (EGY/LEV/ENG) (Baseline) & 81.06 & 80.21 & 71.21 & 67.11 \\ \hline
(1)+ Merged Bilingual CFM & 90.00 & 89.41 & 95.40 & 85.98 \\ \hline
(1)+Merged Bilingual PCS & 89.06 & 88.94 & 94.22 & 83.31 \\ \hline
(1)+Merged Bilingual PseudoCS & 91.96 & 91.92 & \textbf{\textit{96.55}} & \textbf{\textit{86.01}} \\ \hline
(1)+ Merged Multilingual Embeddings(pivot) & 92.81 & 92.91 & 94.81 & 85.87 \\ \hline
(1)+Projected Bilingual & 89.24 & 87.60 & 91.31 & 83.02 \\ \thickhline
 %&  &  &  &  \\ \hline
MTL-POS Tagger (2) + Random-Initi-Embed &  89.91 & 90.02 &  92.89 & 79.51  \\ \hline
%(2) + mono (MSA/SPA/HIN) (baselines) & 85.40 & 83.25 & 50.30 & 63.30 \\ \hline
%(2) + mono (EGY/LEV/ENG) (baselines) & 81.06 & 80.21 & 71.21 & 67.11 \\ \hline
(2)+Merged Bilingual PCS & 90.01 & 90.51 & 93.21 & 84.30 \\ \hline
(2)+Merged Bilingual PseudoCS  & \textit{\textbf{92.90}} & \textit{\textbf{92.92}} & 94.14 & 84.33 \\ \thickhline
MTL-POS+LID Tagger (3) + Random-Initi-Embed & 88.96  & 89.79  & 92.65  & 78.51  \\ \hline
%(3) + mono (MSA/SPA/HIN) (baselines) & 84.11 & 83.23 & 68.21 & 65.02 \\ \hline
%(3) + mono (EGY/LEV/ENG) (baselines) & 80.09 & 79.21 & 49.61 & 62.10 \\ \hline
(3)+ Merged Bilingual CFM & 90.00 & 89.01 & 95.42 & 84.01 \\ \hline
(3)+Merged Bilingual PCS & 90.41 & 90.48 & 95.29 & 84.41 \\ \hline
(3)+Merged Bilingual PseudoCS & 91.89 & 91.92 & 96.50 & 85.91 \\ \hline
(3)+Projected Bilingual & 88.61 & 88.09 & 92.91 & 82.39 \\ \thickhline
State-of-the-art & 90.56 & -- & \textbf{96.63} & \textbf{91.90} \\ \hline
\end{tabular}
\caption{POS tagging accuracy (\%) on the four corpora. Average over five runs with different random seeds. Bold and italics font indicates the best result in our experiments, while bold font indicates the best results compared to the state-of-the-art systems. We refer to BiLSTM-CRF Tagger as (1), MTL-POS Tagger that learns POS tag for related languages as (2), and MTL-POS+LID that learns jointly POS tagging and language identification as (3). Random-Initi-Embed refers to Random initialized embedding}
 
\label{Results-Table}

\end{table*}

%%%%%%%%%%%%%%%%%%%%%%%%%%%%%%%%%%%%%%%%%
%%%%%%%%%%%%%%%%%%%%%%%%%%%%%%%%%%%%%%%%%

%%%%%%%%%%%%%%%%%%%%%%%%%%%%%%%%%%%%%%
%%%%%%%%%%%%%%%%%%%%%%%%%%%%%%%%%%%%%%
%%%%%%%%%%%%Language ID Results%%%%%%%
%%%%%%%%%%%%%%%%%%%%%%%%%%%%%%%%%%%%%%

\begin{table*}[t!]
\small
\center
\renewcommand{\arraystretch}{.90} %to adjust the height of the cells (compress-row)
\begin{tabular}{|c|c|c|c|c|}
\hline
\textbf{Embedding Condition} & \textbf{MSA-EGY} & \textbf{MSA-LEV} & \textbf{SPA-ENG} & \textbf{HIN-ENG} \\ \hline
MTL-POS+LID Tagger: (3) + Random-Initi-Embed & 80.71  & 79.29  & 96.42  & 78.41  \\ \hline

%%%%%%
(3) + mono (MSA/SPA/HIN) (baselines) & 77.41 & 78.08 & 88.11 & 71.46 \\ \hline

(3) + mono (EGY/LEV/ENG) (baselines) & 71.51 & 76.37 & 85.09 & 74.80 \\ \hline
%%%%%%
(3)+ Merged Bilingual CFM & 80.06 & 78.39 & 95.49 & 92.54 \\ \hline
%%%%%%
(3)+Merged Bilingual PCS & 81.33 & 79.28 & 95.02 & 93.20 \\ \hline
%%%%%%
(3)+Merged Bilingual PseudoCS & \textbf{82.15} & \textbf{81.32} & \textbf{\textit{97.20}} & \textbf{\textit{94.92}} \\ \hline
%%%%%%
(3)+Projected Bilingual & 78.17 & 79.11 & 90.01 & 87.69 \\ \thickhline %\hline 
State-of-the-art & -- & -- & \textbf{98.78} & \textbf{97.39} \\ \hline
\end{tabular}
\caption{LID  accuracy (\%) on the four corpora. Average over five runs with different random seeds. Bold and italics font indicates the best result in our experiments, while bold font indicates the best results compared to the state-of-the-art systems. We refer to MTL-POS+LID that learns jointly POS tagging and language identification as (3). Random-Initi-Embed refers to Random initialized embeddings}
 
\label{Results-LID-Table}

\end{table*}
%%%%%%%%%%%%%%%%%%%%%%%%%%%%%%%%%%%%%%
%%%%%%%%%%%%%%%%%%%%%%%%%%%%%%%%%%%%%%

\subsection{Results}
%{\color{red} highlighted}
In this section, we present the results of our experiments using the neural network models and embeddings approach introduced in Section 3 and the datasets from Section 4.1. Also, we show the results of three neural network models when the embeddings are randomly initialized. Table-\ref{Results-Table} shows the POS tagging accuracy results of all language pairs, while Table-\ref{Results-LID-Table} shows the LID accuracy results of all language pairs using MTL-POS+LID Tagger. To evaluate the performance of our approaches we report the accuracy of each condition by comparing the output POS tags generated from each condition against the available gold POS tags for each data set. 
%Moreover, we compare the accuracy of our approaches for each language pair to its corresponding monolingual tagger baseline and the state-of-the-art systems. 
We consistently apply the different experimental conditions on the same test set per language pair: for MSA-EGY we report results on MSA-EGY test set, for MSA-LEV we report results on MSA-LEV test set, and for SPA-ENG, we report results on Miami Bangor corpus (SPA-ENG) test set, and finally for HIN-ENG, we report on UD-HIN-ENG (HIN-ENG) test set. 
The highest accuracy results for MSA-EGY and MSA-LEV language pairs are 92.90\% and 92.92\%, respectively. 
These results are achieved by MTL-POS Tagger+Merged Bilingual PseudoCS embeddings. Our best model for MSA-EGY outperform state-of-the-art system by $\sim$ 2\% \cite{MyPOSPaper}. We could not compare our best system for MSA-LEV language pair to any previous systems as we map the original POS tag set (Buckwalter POS tag set) of the MSA-LEV dataset into UD POS tag set.  
%We could not compare our best system as our work is the first study that uses MSA-LEV data set with Universal POS tag set \cite{UDPOS}. 
BiLSTM-CRF+Merged Bilingual PseudoCS embeddings yield the highest accuracy results for both SPA-ENG and HIN-ENG language pairs. The SPA-ENG model's accuracy, 96.55\%, is comparable to the state-of-the-art system, 96.63\% \cite{VictorPaper}. On the other hand, the accuracy of our best HIN-ENG model (86.01\%) underperforms the state-of-the-art system (91.90\%) \cite{bhat2018universal}. 
For LID task, all our models underperform the state-of-the-art systems. Since there are no state-of-the-art systems for MSA-EGY and MSA-LEV, we compare the performance of our models against the baseline systems. 
%The MTL-POS+LID Tagger+Merged Bilingual PseudoCS model outperform all other experimental setups across the board for all language pairs. 

%All models that utilize cross-lingual embeddings outperform the other neural network architectures that use the various proposed embeddings setups. The second highest accuracy results go to the models that use CS embeddings for MSA-EGY, MSA-LEV and HIN-ENG, and to the models that use CS embeddings for SPA-ENG. 
%For SPA-ENG, the BiLSTM-CRF that used Cross-lingual embeddings reaches the best accuracy, 96.91\%. For HIN-ENG language pair, the Combined:Mono embeddings yields the best accuracy, 95\%.

\section{Discussion}
%Multilingual embedding 
Multilingual embedding (e.g., MSA-EGY and MSA-LEV) helps closely related languages (EGY/LEV) but adds noise to the languages that are distant (SPA/HIN). Similarly, learning jointly POS tagging for closely related languages yields the highest accuracy results for MSA-EGY and MSA-LEV as opposed to the languages that are distant (SPA/HIN). The accuracy results of MSA-EGY and MSA-LEV language pairs are the highest results in all experimental setups. The improvement could be attributed to the significant number of homographs some of which are cognates. 

%%%%%%%%%%%%%%%%%%%%%
%%%%%%%%%%%%%%%%%%%%%%%%%%%%%%%%%%%%%%%%%%%%%
%%%%%%%%%%%%%%%Error Rate Table%%%%%%%%%%%%%%
%%%%%%%%%%%%%%%%%%%%%%%%%%%%%%%%%%%%%%%%%%%%%
\begin{table*}[t!]
\renewcommand{\arraystretch}{.75} %to adjust the height of the cells (compress-row)
\setlength{\tabcolsep}{0.2em}
\small
\begin{center}
%\resizebox{0.5}
%\resizebox{\columnwidth}{!}{%
\begin{tabular}{|l|c|l|l|c|l|l|c|l|l|c|}
\cline{1-2} \cline{4-5} \cline{7-8} \cline{10-11}
\textbf{MSA-EGY} & \multicolumn{1}{l|}{\textbf{}} & \textbf{} & \textbf{MSA-LEV} & \multicolumn{1}{l|}{\textbf{}} & \textbf{} & \textbf{SPA-ENG} & \multicolumn{1}{l|}{\textbf{}} & \textbf{} & \textbf{HIN-ENG} & \multicolumn{1}{l|}{} \\ \cline{1-2} \cline{4-5} \cline{7-8} \cline{10-11} 
\multicolumn{1}{|c|}{\textbf{Error Type}} & \textbf{Percentage} & \multicolumn{1}{c|}{\textbf{}} & \multicolumn{1}{c|}{\textbf{Error Type}} & \textbf{Percentage} & \multicolumn{1}{c|}{\textbf{}} & \multicolumn{1}{c|}{\textbf{Error Type}} & \textbf{Percentage} & \multicolumn{1}{c|}{\textbf{}} & \multicolumn{1}{c|}{\textbf{Error Type}} & \textbf{Percentage} \\ \cline{1-2} \cline{4-5} \cline{7-8} \cline{10-11} 

ADJ \textgreater NOUN & 19\% &  & ADJ \textgreater NOUN & 21\% &  & NOUN \textgreater ADJ & 10\% &  & NOUN \textgreater VERB & 27\% \\ \cline{1-2} \cline{4-5} \cline{7-8} \cline{10-11} 

VERB \textgreater NOUN & 15\% &  & VERB \textgreater NOUN & 19\% &  & NOUN \textgreater PRON & 8\% &  & VERB \textgreater NOUN & 24\% \\ \cline{1-2} \cline{4-5} \cline{7-8} \cline{10-11} 

NOUN \textgreater VERB & 11\% &  & NOUN \textgreater ADJ & 16\% &  & VERB \textgreater NOUN & 7\% &  & NOUN \textgreater ADJ & 19\% \\ \cline{1-2} \cline{4-5} \cline{7-8} \cline{10-11} 

NOUN \textgreater ADJ & 8\% &  & NOUN \textgreater VERB & 9\% &  & ADJ \textgreater PRON & 5\% &  & ADJ \textgreater VERB & 14\% \\ \cline{1-2} \cline{4-5} \cline{7-8} \cline{10-11} 

%PART \textgreater NOUN & 7\% &  & PART \textgreater NOUN & 7\% &  & PRONP \textgreater NOUN & 4\% &  & NOUN \textgreater VERB & 9\% \\ \cline{1-2} \cline{4-5} \cline{7-8} \cline{10-11} 
\end{tabular}
%}
\end{center}
\caption{Most common errors for the best systems for all language pairs (Gold-POS $>$ Predicted-POS)}
\label{Error-rate-table}
\end{table*}
%%%%%%%%%%%%%%%%%%%%%%%%%%%%%%%%%%%%%%%%%%%%%
%%%%%%%%%%%%%%%%%%%%%%%%%%%%%%%%%%%%%%%%%%%%%
%%%%%%%%%%%%%%%%%%%%%%%%%%%%%%%%%%%%%%%%%%%%%
The CS behavior can be different depending on the medium of communication, topic, speakers (or authors), and the languages being mixed among other factors. Hence, we believe that the difference in the genre of the evaluation data sets of SPA-ENG and HIN-ENG language pairs is one of the potential reasons that make both language pairs not to benefit from the multilingual embedding and learning jointly POS tagging for both language pairs.  
On the other hand, the MTL-POS-LID Tagger that learns simultaneously POS and language id tags with the aim of boosting the performance of POS tagging task benefit distant (SPA/HIN) more than closely related languages (EGY/LEV). 
%%%%%%%%%%%%%%%%%%%%%%%%%%%%%%%%%%%%%%%
%%%%%%%%%%%%%%%%%%%%%%%%%%%%%%%%%%%%%%%
%The code-switching behavior can be different depending on the medium of communication, context of language use, topic, authors (or speakers), and the languages being mixed among other factors.
%%%%%%%%%%%
We define code switching points as the points within a sentence where the languages of the words on the two sides are different. We observe a sharp jump in the accuracy for SPA-ENG corpus. We believe the major factor of this jump is the low percentage of the CS points, $\sim$ 6\%, while the percentage of CS points in the MSA-EGY, MSA-LEV, and HIN-ENG datasets are relatively high, 38.78\%, 30.12\%, and 15.17\%. The low percentage of CS points in the SPA-ENG corpus leads the models that address the inter-sentential code switching type (BiLSTM-CRF+Merged BilingualMerged Bilingual CFM and MTL+Merged BilingualMerged Bilingual CFM) to score the second highest accuracy results, 95.40\% and 95.42\%. 

The two key advantages of the Merged Bilingual PseudoCS embeddings and Multilingual embeddings  are, 1) it enables the learned embeddings to capture the interactions between the words in different languages; 2) It captures the word usage in the context of each language and eliminates the ambiguity for the words that have the same surface form in multiple languages. Hence, the OOV percentage and ambiguity of words are reduced. Using multilingual embeddings for MSA-EGY and MSA-LEV, reduced the percentage of OOV of MSA-EGY and MSA-LEV from 10\% and 13\% to 8\% and 10\%, respectively. Similarly, with SPA-ENG and HIN-ENG language pairs, the rate of OOV is decreased from 12\% and 15\% to 9\% and 11\%, respectively.
%%%%%%%%%%%%%%%%%%%%%
%%%%% Review %%%%%%%%
%%%%%%%%%%%%%%%%%%%%%
%Code- Switching pattern

One of the common CS intra-sentential patterns we notice in our data sets is insertion patterns. This pattern involves inserting material (lexical items, or entire constituents) from one language into a structure from the other language \cite{muysken}. 
To evaluate the effect of the CS insertion pattern we define the CS fragment (CSF) of those test sentences. We define a CSF as the minimum contiguous span of words where a CS occurs \cite{VictorPaper}. 
%For HIN-ENG and SPA-ENG, the length of the CS fragment is ranged from two to three words long, spanning a Spanish(or Hindi) token and an English one or vice versa but it can be longer.
The average length of the CS fragments in the SPA-ENG test set is 2.16, and 6.1 in the HIN-ENG test set. 
%For MSA-EGY and MSA-LEV, the length of the CS fragment is ranged from four to five words. 
The average length of the CS fragments is 5.1 in MSA-LEV, and 5.8 in MSA-EGY test set. 
We observe that the length of the CS fragments impacts the overall performance of the classifiers. For example, short CS fragments confuse the best classifiers of almost all language pairs. We noticed that a majority of CS sentences that have one or two lexical elements inserted had been miss-classified by almost all models in all language pairs. 
%%%%%%%%%%%%%%%%%%%%%%%%%%%%
%Bilingual embeddings

Using bilingual embeddings outperform the baseline systems, but it did not achieve the highest results achieved by the other proposed models for all language pairs. The experiment results show a promising direction towards obtaining bilingual embeddings for CS tasks. Our explanation for this performance is that the sense distribution of polysemous words can differ widely between a monolingual (mono) corpus and Merged Bilingual PseudoCS corpus. For instance, the word ‘bank’ in English has several meanings such as (a) the land alongside or sloping, (b) a financial institution, and (c) a set or series of similar things. However, in a Spanish dominant sentence or corpus, 'bank' is primarily, if not only, used in sense (a). Our experiments show that standard bilingual embeddings are not well suited, in general, for CS tasks; embeddings learned from CS data yield better results which are aligned with our findings \cite{D18-1344}. 
%We expect that pretrained bilingual word embeddings can help in improving the performance POS tagging models, especially for MSA-LEV and MSA-EGY as they are closely related with many cognates. However, the results achieved show the apposite. Thus, our experiments shows that standard bilingual embeddings are not well suited, in general, for CS tasks; embeddings learned from CS data yield better results our finding is aligned with \cite{D18-1344} 
%Error Rate

Table \ref{Error-rate-table} shows the most common errors for the best systems for each language pairs. We observe almost the same trends across both MSA-EGY and MSA-LEV language pairs, while the common errors are relatively different between SPA-ENG and HIN-ENG.    

\section{Conclusion}

%In this paper, we have presented a comparison of multiple approaches for POS tagging of CS data in three language pairs. Our evaluation shows bilingual embedding benefit languages/dialects that share very close linguistic roots. In addition, we studied the impact of using various set-ups for training word embedding including bilingual embedding, monolingual embedding, CS embedding as well as learning POS tags jointly with language id tags, on the performance of POS tagger.
In this paper, we present a detailed study of various strategies for POS tagging of CS data in four language pairs. We explore multiple strategies of measuring the impact of pretrained embeddings on POS tagging of CS data. We find that related language pairs, e.g., MSA-EGY and MSA-LEV, benefit from both jointly learning POS tagging as well as merged multilingual embeddings (i.e., pivot embedding), while distant language pairs, e.g., SPA-ENG and HIN-ENG, benefit from a multi-task learning model that learns two different tasks, e.g., POS tagging and language identification.  Furthermore, we compared our results to the previous state-of-the-art POS tagger for MSA-EGY, SPA-ENG, and HIN-ENG and showed that our classifiers outperform the MSA-EGY state-of-the-art system in every configuration \cite{MyPOSPaper}. The results achieved by BiLSTM-CRF+Merged Bilingual PseudoCS embeddings model is comparable to \newcite{VictorPaper}.  We will explore several directions in the future. First, we will study the theoretical aspects of word embedding learning. Second, we will investigate the proposed word embeddings on other downstream NLP applications, such as segmentation and parsing. 

%%%%%%%%%%%%%%%%%%%%%%%%%%%%%%%%%%%%%%%%%%%%%%%%%%%%%%%%%%%%%%%%%%%
%%%%%%%%%%%%%%%%%%%%%%%%%%%%%%%%%%%%%%%%%%%%%%%%%%%%%%%%%%%%%%%%%%%
\section*{Acknowledgments}
We would like to thank the four anonymous reviewers for their valuable comments and suggestions.

%\noindent {\bf Preparing References:} \\
%Include your own bib file like this:
%\verb|\bibliographystyle{acl_natbib}|
%\verb|\bibliography{naaclhlt2019}| 

%where \verb|naaclhlt2019| corresponds to a naaclhlt2019.bib file.
\bibliography{naaclhlt2019}
\bibliographystyle{acl_natbib}

\appendix

\end{document}